\documentclass{article}

\usepackage[preprint]{neurips_2024}

\usepackage[utf8]{inputenc} % allow utf-8 input
\usepackage[T1]{fontenc}    % use 8-bit T1 fonts
\usepackage{hyperref}       % hyperlinks
\usepackage{url}            % simple URL typesetting
\usepackage{booktabs}       % professional-quality tables
\usepackage{amsfonts}       % blackboard math symbols
\usepackage{nicefrac}       % compact symbols for 1/2, etc.
\usepackage{microtype}      % microtypography
\usepackage{xcolor}         % colors

\usepackage{amsmath,amsfonts,bm}

\def\eqref#1{equation~\ref{#1}}
\def\1{\bm{1}}

\def\rvr{{\mathbf{r}}}
\def\rvs{{\mathbf{s}}}

\def\rvx{{\mathbf{x}}}

\def\rvz{{\mathbf{z}}}

\DeclareMathAlphabet{\mathsfit}{\encodingdefault}{\sfdefault}{m}{sl}
\SetMathAlphabet{\mathsfit}{bold}{\encodingdefault}{\sfdefault}{bx}{n}

\usepackage{amsmath}
\usepackage{amssymb}
\usepackage{mathtools}
\usepackage{amsthm}

\usepackage{amssymb}
\usepackage{xcolor}
\usepackage{wrapfig}

\newcommand{\mysquare}[1][black]{\textcolor{#1}{\ensuremath\blacksquare}}

\usepackage[capitalize,noabbrev]{cleveref}

\theoremstyle{plain}
\newtheorem{theorem}{Theorem}[section]

\theoremstyle{definition}

\theoremstyle{remark}
\newtheorem{remark}[theorem]{Remark}

\newlength\savewidth
\usepackage{xcolor}
\usepackage{tabulary}
\newcolumntype{x}[1]{>{\centering\arraybackslash}p{#1pt}}

\title{Video Occupancy Models}

\author{
    Manan Tomar$^1$ ~
    Philippe Hansen-Estruch$^2$ ~ 
    Philip Bachman ~ 
    Alex Lamb$^3$ ~\\
    \And John Langford$^3$ ~
    Matthew E. Taylor$^1$ ~
    Sergey Levine$^4$ ~ \\ \\
    University of Alberta$^1$, \ \ UT Austin$^2$, \ \ Microsoft Research NYC$^3$, \ \ UC Berkeley$^4$ \\
    \texttt{\{manan.tomar\}@gmail.com}
}

\begin{document}

\maketitle

\begin{abstract}
We introduce a new family of video prediction models designed to support downstream control tasks. We call these models Video Occupancy models (VOCs). VOCs operate in a compact latent space, thus avoiding the need to make predictions about individual pixels. Unlike prior latent-space world models, VOCs directly predict the discounted distribution of future states in a single step, thus avoiding the need for multistep roll-outs. We show that both properties are beneficial when building predictive models of video for use in downstream control. Code is available at \href{https://github.com/manantomar/video-occupancy-models}{\texttt{github.com/manantomar/video-occupancy-models}}.
\end{abstract}

\section{Introduction}

Modeling the future is essential for planning. Predicting future events has long been a fundamental principle for learning in animals~\citep{stachenfeld2017hippocampus}, driving recent deep learning research to focus on building better predictive models~\citep{oord2018representation, ho2022imagen}. Modeling the future from video data presents a significant challenge that requires addressing two fundamental questions. First, how detailed should such a model be? Should it produce pixel-level predictions, or should it operate at a more abstract level, such as a latent representation of the input? Second, how far into the future should the model predict? Should sampling from the model depend on a specific temporal timestep? This question is inherently linked to how many video frames a model processes at a given time to produce future samples. In the context of planning for downstream control, these remain open questions with multiple valid answers and various trade-offs.

Regarding the first question, making predictions directly in the pixel space of input images is computationally complex and costly, often expending resources on predicting elements that are not useful for control. Capturing the essential information in a compact latent space and making predictions within this space is frequently a more efficient alternative. Significant progress has been made in learning better latent representations via self-supervised learning~\citep{he2022masked}, and powerful generative models have been developed to make predictions in these latent spaces. These techniques offer good solutions to capturing information more abstractly but have seen limited use in making temporal predictions.

For the second question, most predictive models typically learn to make one-step predictions, which can be combined autoregressively to produce predictions further into the future. The successor representation (SR)\citep{dayan1993improving} captures the expected occupancy of future states in a single representation but does not provide the full sampling capability of a dynamics model. Recently, a generative analog of the SR, called $\gamma$-models\citep{janner2020gamma}, was introduced. $\gamma$-models can sample from the discounted future state occupancy distribution, reducing the need to unroll a standard one-step model over multiple timesteps. However, $\gamma$-models have only been applied to problems with small state spaces and have not addressed the challenges of learning compact representations from rich pixel-level video data.

\begin{figure*}[t]
    \centering
    \includegraphics[trim={5cm 11cm 7cm 10cm}, clip, width=1\linewidth]{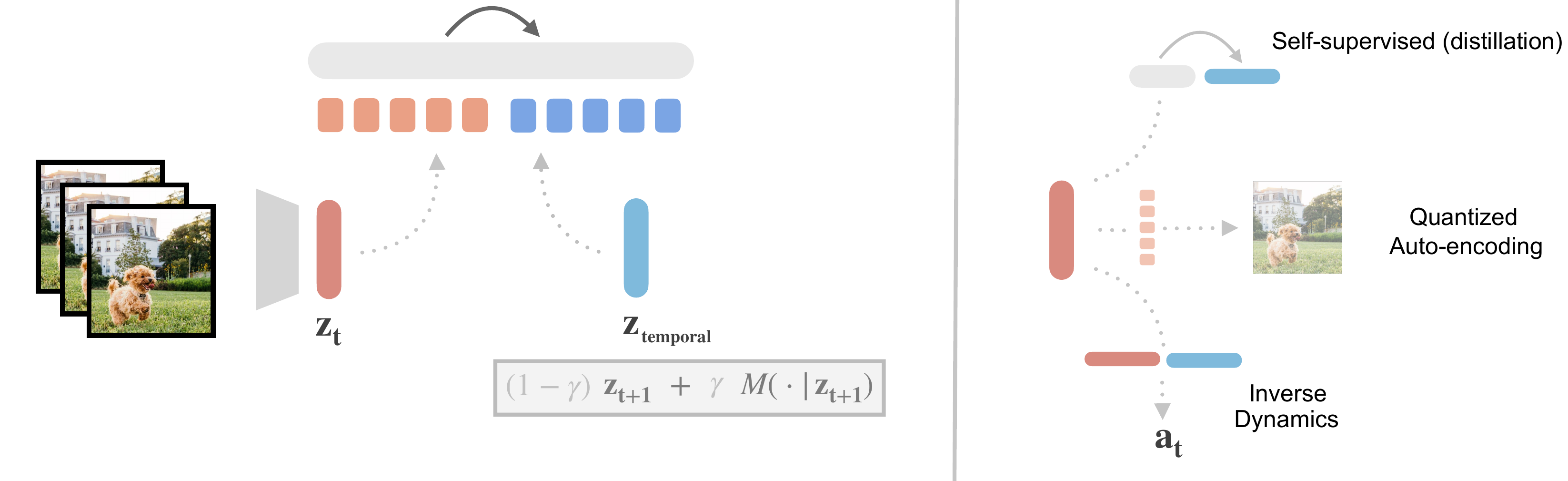} %{left bottom right top}
    % \vspace{2mm}
    \caption{\textbf{Video Occupancy Models}. \textit{Left.} A small stack of pixel observations is encoded in a representation $\rvz_t$, which is then quantized to produce discrete tokens for a GPT model \big(denoted by $M(\cdot | \rvz_{t})$\big). A \textit{temporal target} is encoded in a similar way by sampling the next representation $\rvz_{t+1}$ with probability $(1 - \gamma)$ or a bootstrap sample from the model conditioned on $\rvz_{t+1}$ with probability $\gamma$. The GPT model then does next token prediction on the concatenated tokens corresponding to the current representation (red tokens) and the temporal target (blue tokens). \textit{Right.} The representation $\rvz_t$ is learnt in three different ways, including 1) quantized autoencoding such as VQ-VAE, 2) inverse dynamics modelling in the presence of action data, and 3) a self-supervised distillation based objective that gives away doing pixel-level reconstruction in favor of learning to predict in a latent space.}
    \label{fig:main}
\end{figure*}

This paper brings together the best aspects of these ideas. We propose a generative model that directly predicts the discounted future distribution of representations of observations. We employ generative temporal difference (TD) learning~\citep{sutton1988learning} to predict future representations. In particular, we compute a generative target for the current observation by sampling the representation of the next observation with probability $1 - \gamma$ and sampling a future representation from the model bootstrap with probability $\gamma$. Subsequently, we implement this generative TD algorithm using an autoregressive transformer architecture over a latent representation space. We explore transforming raw observations into latent representations through three methods: 1) quantized autoencoding (VQ-VAEs)\citep{van2017neural}, 2) inverse dynamics modeling\citep{lamb2022guaranteed}, which utilizes action data (if present), and 3) a self-supervised distillation based objective~\citep{caron2021emerging} that avoids pixel-level prediction in favor of learning to predict in a latent space\footnote{To the best of our knowledge, this is the first world model that has been shown to work with self-supervised representations such as DINO.}. We refer to our approach as Video Occupancy Models (VOCs). 

Like the successor representation, VOCs capture temporal information about the environment dynamics. However, VOCs learn the full discounted future (latent) state distribution rather than just the expected value of this distribution. Unlike previous generative models applied in control, VOCs work over a representation space, one that is learnt via various self-supervised methods, i.e. without explicit labels or reward information, and is thus able to work well over high-dimensional pixel spaces. Finally, unlike standard representation learning models which output a single feature vector, our model can produce multiple future predictions for any given inferred representation in the video.

We emphasize two key motivations for Video Occupancy Models: 1) avoiding pixel-level predictions, and 2) not predicting at every temporal time-step. We demonstrate that models which showcase these properties remain effective for downstream control, while leading to improved and faster prediction capabilities. Our contributions are: 1) a family of generative models called VOCs that operate on latent spaces and are trained via Temporal Difference learning, 2) demonstrating how VOCs can be used for efficient value estimation, and 3) incorporating VOC-based value estimation in a model predictive control framework for downstream control.

\section{Background}

\subsection{Successor Representation and Gamma Models}

The successor representation (SR)~\citep{dayan1993improving} aims to capture a summary of future state occupancy. Specifically, it learns to estimate the expected occupancy of visiting a future state $\rvs_e$ and is denoted as $M(\rvs_e \ | \ \rvs_t)$, where the conditioning is on the current state $\rvs_t$. The SR can be learnt similar to how value functions are learnt, via temporal difference (TD) backups, described as follows:

\vspace{-.5cm}
\begin{equation}
    M(\rvs_e \ | \ \rvs_t) = \mathbb{E}_{\rvs_{t+1}} \Big(\underbrace{\mathbb{I} [\rvs_e = \rvs_{t+1}]}_{\text{SR reward}} + \gamma M(\rvs_e \ | \ \rvs_{t+1})\Big),
\end{equation}

where the reward is replaced by an indicator, SR reward. The SR reward is 1 if the next state is the same as the state for which the SR is computed, and 0 otherwise. When the state space is high dimensional, such an indicator reward can be replaced by a continuous \textit{feature reward}, in which case the SR equivalently captures future \textit{feature} or \textit{representation} occupancy. The feature reward essentially refers to the features associated with an observation $\rvx_t$, denoted as $\phi(\rvx_t)$. We call the features as a reward since they substitute for the reward in the TD update. This version of the SR is then fittingly termed Successor Features~\citep{kulkarni2016deep}.

\begin{equation}
    M(\phi(\rvx_e) \ | \ \phi(\rvx_t)) = \mathbb{E}_{\rvs_{t+1}} \Big(\underbrace{\phi(\rvx_t)}_{\text{SF reward}} + \gamma M(\phi(\rvx_e) \ | \ \phi(\rvx_{t+1}))\Big),
\end{equation}

Instead of learning to estimate the \textit{expected} state occupancy, one could also learn to estimate the entire distribution. Gamma models~\citep{janner2020gamma} achieves this capability via generative TD based sampling and uses either GANs or Flow models for capturing the SR distribution. In the next section, we will utilize the same idea of generative TD learning for learning distributions of future representations of observations.

\subsection{Quantized AutoEncoding and Self-Supervised Models}

Vector Quantized Variational Autoencoders (VQ-VAEs)~\citep{van2017neural} are a type of generative model designed to create high-quality discrete latent representations. They consist of an encoder that transforms input data into a latent space, a quantizer that maps these continuous latent vectors to a finite set of vector embeddings (codebook), and a decoder that reconstructs the data from these quantized vectors. 

DINO~\citep{caron2021emerging} (self-Distillation with No labels) is a self-supervised learning method aimed at learning visual representations without labeled data. It employs student and teacher networks of identical architecture, and uses the teacher network's predictions to align the student network's predictions. DINO uses stop-gradient and exponential moving average (EMA) on the teacher network to avoid representation collapse. This approach enables DINO to learn robust and transferable visual representations from large-scale unlabeled datasets, demonstrating high performance in various downstream tasks such as image classification and object detection.

\section{Video Occupancy Models}

\begin{figure*}[t]
    \centering
    % left bottom right top
    % \includegraphics[trim={13cm 15cm 20cm 10cm}, clip, width=\linewidth]{Figures/gamma_variations_2.pdf}
    \includegraphics[trim={0cm 0cm 0cm 0cm}, clip, width=\linewidth]{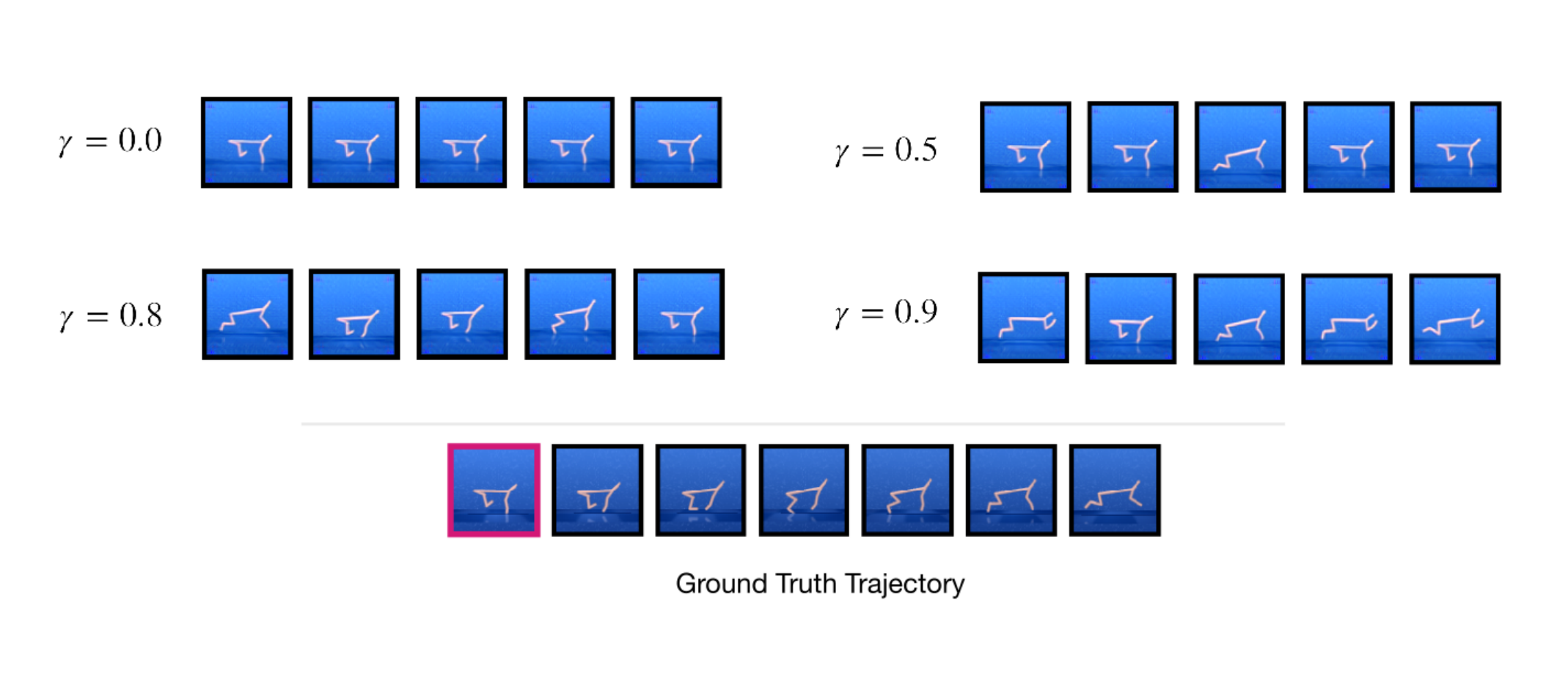}
    \vspace{-6mm}
    \caption{\textbf{Gamma Variations}. Visualization of predictions made by Video Occupancy Models (VOCs) with a VQ-VAE representation space, for different $\gamma$ values. The predictions are made in the latent space and then decoded via the VQ-VAE decoder. The bottom row shows the ground truth trajectory, with the observation highlighted with \mysquare[magenta] (magenta) denoting the conditioning state for the VOC. Each row for a given $\gamma$ value includes five independent samples produced by the VOC. All predictions are based on a single forward pass through the model. As the $\gamma$ value increases, the model produces longer term predictions within a single step. For $\gamma=0$, we recover a standard 1-step model, with all predictions being identical to the ground truth next state. For high gamma values (e.g. $\gamma = 0.9$), the predictions are less similar to the ground truth observations, since the model is asked to produce long term predictions.}
    \label{fig:gamma-variations}
\end{figure*}

Video Occupancy models consist of two parts, a dedicated \textit{representation space} which captures information present in a short sequence of video frames, and a \textit{generative model} that can produce temporal predictions over the representation space. A natural choice for the generative model is to use an autoregressive model and so we exclusively use a GPT-2~\citep{radford2019language} model that autoregressively predicts a sequence of tokens. The representation space is responsible for producing this sequence of tokens that get fed into the generative model (see Figure~\ref{fig:main}). In particular, a continuous representation of a sequence of frames/observations is quantized into discrete tokens that are used as input to the generative model. Instead of processing a long sequence of frames, VOCs only encode a small number of observations into the current representation $\rvz_t$. A \textit{temporal target} is then computed in a similar fashion, capturing information present in future observations. The quantized versions of the current and the temporal target representation are then concatenated to form a sequence of discrete tokens. The GPT model (denoted as $M$) is trained to do next token prediction on this concatenated sequence of tokens. The temporal target is computed such that it captures future information within a single representation time-step. To achieve this, we sample the temporal target from a mixture distribution, that between the next time-step (quantized) representation $\rvz_{t+1}$ and a bootstrapped output of the GPT model conditioned on the next time-step representation $M(\cdot | \rvz_{t+1})$. The temporal target $\rvz_{\text{temporal}}$ can thus be defined as a sample from the following mixture distribution: $(1 - \gamma) \ p(\cdot \ | \ \rvz_t) + \gamma M'(\cdot \ | \ \rvz_{t+1})$, where the $M'$ is the bootstrapped version of the GPT model. The sampling between these two distributions within the mixture is governed by the parameter $\gamma$, which describes how myopic or farther-away-in-time the temporal target should be. In the case that $\gamma = 0$ for instance, all the sampling weight is put on the next time-step distribution and so we recover a standard 1-step model. In all other cases, the bootstrap model output plays a role in sampling the temporal target and hence it carries more future information than a 1-step target. Note that this way of computing the temporal target is exactly how TD-learning functions, where the next time-step representation replaces the reward, and the parameter $\gamma$ is typically called the discount factor. Once such a model is learnt, a single forward pass will generate a future representation (governed by the value of $\gamma$), circumventing the issue of having to do multiple forward passes through a one step model to produce a future prediction.

With the TD-learning perspective in mind, it is fairly straightforward to view VOCs as a highly expressive generative analog of the successor representation (SR). In the SR context, future predictions are typically learnt by capturing an expectation of future state occupancy via TD learning using an indicator reward. However, when the observations are high dimensional, comparing observations pixel-by-pixel is clearly not a good idea in order to capture similarity. To that end, VOCs allow for learning a useful representation that only capture information about the abstract structure present in the observations, and not focus on the lower level detail, such as pixel intensities. Additionally, capturing an expectation of the future representations (as is typically done for learning the SR) is too limited as it does not allow us to use the model to sample future observations. An analogous generative version switches to learning the distribution of occupancy of future observations rather than just the expectation and thus allows for sampling from the model. These two properties produce a powerful combination within a single model.

Having described how the generative component of VOCs work, we now focus on how the representations are learnt. In this paper, we describe three different ways of doing so, each leading to a VOC model with different properties. The first method is the most straightforward, that of quantized autoencoding, which includes VQ-VAEs. VQ-VAEs have been used extensively for tokenization of images, particularly because they preserve a lot of pixel information within the learnt representation, thus leading to strong long term predictions within a temporal model that processes the corresponding tokens. The second method utilizes inverse dynamics modelling, in the case there are actions present alongside video data. Learning a multi-step inverse dynamics model has been shown to capture long-term dependencies in its representation~\citep{lamb2022guaranteed}. Finally, as the third method, we use the self-supervised DINO method~\citep{caron2021emerging} to learn representations, and follow it with a quantization step to produce corresponding tokens for the generative model.

\textbf{Value Estimation with VOCs.} One of the advantages of learning VOCs is that we can use them as a drop-in replacement for the successor representation, which can be used for estimating value functions. Since both the SR and VOCs only capture dynamics information, we must separately learn a reward model to compute value estimates. We learn such a reward model over the VOC representation space instead of over the pixel space. In what follows, we describe two ways to estimate value functions using a learned reward model. Since the temporal model is autoregressive over a discrete input space, it can produce high quality samples as well as density estimates, and thus we can estimate value functions through either method:

\textit{Value Estimation via Sample Generation}. Since VOCs capture future occupancy of observations, we can simply sample from the model, and compute the average reward obtained for the sampled states w.r.t to the reward model. More concretely, we train a reward model $\rvr(\cdot)$ on top of the representation space, which is then used to compute value estimates as follows:

\begin{equation}
\label{eq:value-estimation-sample}
    V(\rvs) = \frac{1}{1 - \gamma} \ \mathbb{E}_{\rvs_e \sim M} \ \rvr(\rvs_e),
\end{equation}

where $\rvs_e$ are different samples generated from a VOC model.

\textit{Value Estimation via Density Evaluation}. Another alternative to estimating values is by querying VOCs for the probability density of a target sequence of future observations and then computing their sum, as follows:

\begin{equation}
\label{eq:value-estimation-density}
    V(\rvs) = \sum_{t=0} p_{M}(\rvs_t) \ \rvr(\rvs_t),
\end{equation}

where $p(\rvs_e)$ is the corresponding density estimates of states $\rvs_e$ as estimated by the VOC model. 

Note that the $\gamma$ discounting is taken care of directly in the training of VOCs through the $\gamma$ based sampling of target representations. Therefore, simply sampling from the model or computing density estimates directly gives us value estimates.

\section{Experimental Analysis}
In this section, we discuss three different instantiations of VOCs depending on how the representation space is constructed, namely -- quantized autoencoding, inverse dynamics modelling, and self-supervised distillation. Additional details including hyperparameters are included in the Appendix. Finally, we include results on using VOCs in a model predictive control framework.

\subsection{Quantized AutoEncoding}

\textbf{VQ-VAE VOCs.} We begin by learning Video Occupancy models with a VQ-VAE representation space. We take a VQ-VAE model pretrained on ImageNet and then finetune it for specific MuJoCo-based datasets~\citep{todorov2012mujoco, tassa2018deepmind}. The discrete codes produced by the VQ model act as tokens for the pixel-based observations. Each VQ encoded observation produces a 25-dim vector of discrete values with a vocabulary size of 1024. We then concatenate the VQ codes for three consecutive observations to produce the conditioning state $\rvz_t$, akin to how frame stacking is necessary to capture and predict velocity information. We similarly generate a 75-dim (25-dim code vector $\times$ 3 observations) temporal target $\rvz_{\text{temporal}}$ and then concatenate the conditioning state and the target to form the input ($\{\rvz_t, \rvz_{\text{temporal}} \}$) to the GPT-2 model. The generative model then does next token prediction on this input. The bootstrapped target for the temporal difference (TD) update is generated by multinomial sampling from the GPT model (we do not use beam based sampling or greedy sampling strategies for generating the target). Note that temporal prediction is strictly limited to predicting what is captured in the representation space and is not driven by the pixel-level reconstruction loss. Since the representation is defined via a VQ-VAE, at inference we can sample future representations from the learnt model and use the the VAE's decoder to obtain pixel-level predictions, providing a natural way to inspect prediction quality. Figure~\ref{fig:gamma-variations} shows sampled outputs from VOCs trained for different $\gamma$ values. For $\gamma = 0$, we recover a single step model. As the $\gamma$ value increases, the model makes longer-in-time predictions.

\begin{figure*}[t]
    \centering
    % left bottom right top
    \includegraphics[trim={5cm 15cm 5cm 2cm}, clip, width=\linewidth]{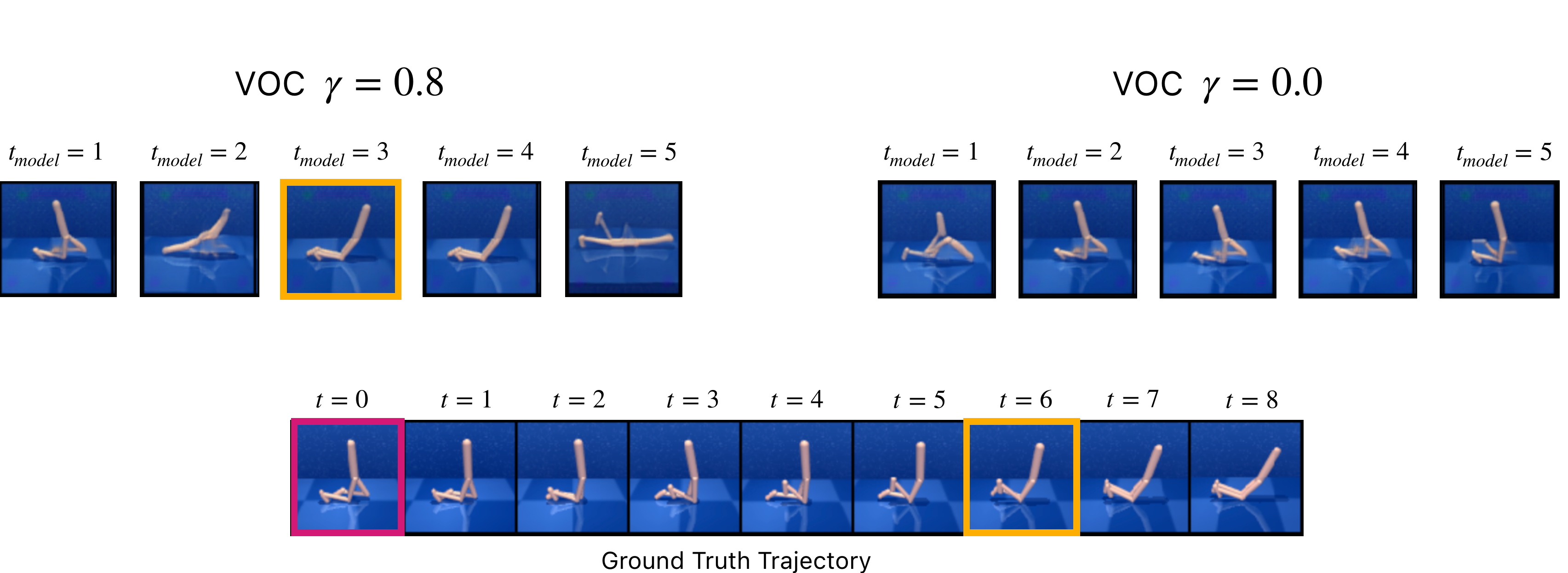}
    \vspace{-4mm}
    \caption{\textbf{Model Rollouts}. for Video Occupancy Models with $\gamma=0.8$ (\textit{top-left}) and 1-step model, i.e. $\gamma=0.0$ (\textit{top-right}). Both models are learnt over the same VQ-VAE representation space and are conditioned on the first observation (shown in magenta) in the ground truth trajectory (bottom row). $t$ refers to the timestep in the ground truth trajectory, while $t_{model}$ refers to the number of forward passes made by the model. For $\gamma=0.8$, a single sample from the model ($t_{model} = 1$) yields a farther-in-time ($t \ge 1$) prediction. For instance, the frame marked with yellow shows how the VOC model can predict the $t=6$ observation within $t_{model}=3$ steps. On the other hand, a 1-step model must be unrolled for multiple timesteps autoregressively to obtain a prediction a future prediction.}
    \label{fig:rollouts}
\end{figure*}

\begin{figure}[t] %{r}{0.5\textwidth}
    \centering
    % left bottom right top
    % \vspace{-.2cm}
    \includegraphics[trim={0cm 16cm 2cm 3cm}, clip, width=\linewidth]{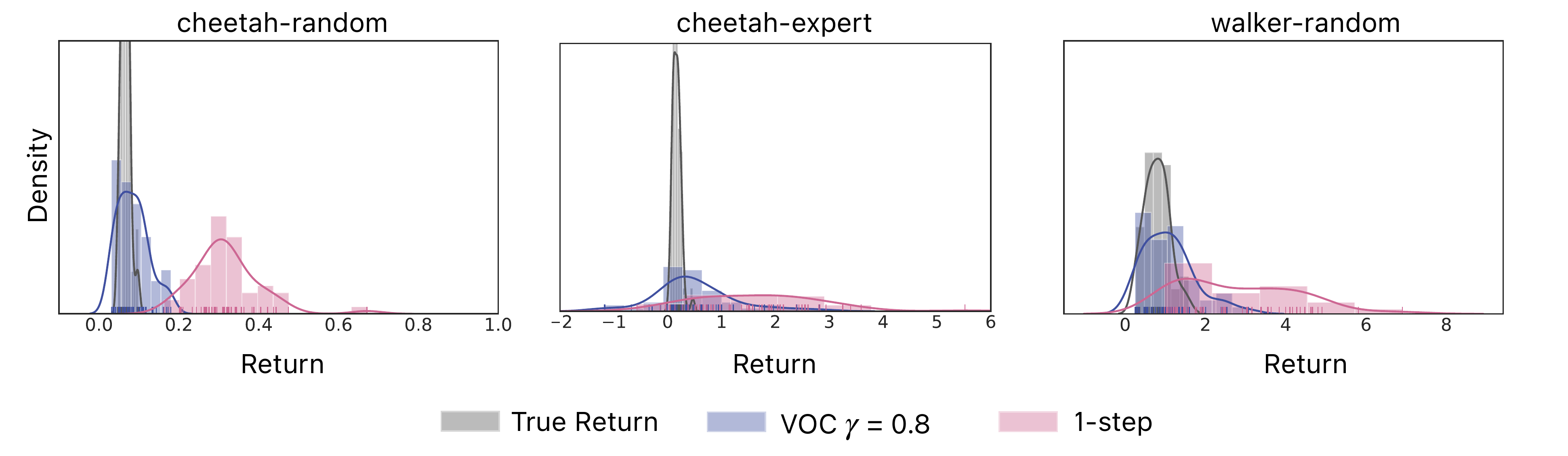}
    \caption{\textbf{Return Distribution Estimation with VOCs}. We train a reward model on the VQ representation space pre-discretization (reward loss is shown in the top row) and then use the learned reward model to plot the return distribution of a state by sampling from a Video Occupancy Model, and compare it with a one-step Model (Eq.~\ref{eq:value-estimation-sample}).}
    \label{fig:pgm-value}
\end{figure}

\vspace{0.5cm}
\textbf{Benefit over a 1-step model.} We compare VOCs with a standard one-step model, which essentially corresponds to setting $\gamma=0$ for the VOC. Note that a single rollout step of the VOC models corresponds to a multi-step prediction, whereas a 1-step model must be unrolled for longer to produce a prediction corresponding to a given horizon length. A 1-step model thus suffers from accumulating errors in the autoregressive prediction the most. A VOC model on the other hand remedies this issue by using the TD-based backups in its model training and thus requires only a single pass to produce a multi-step prediction. The difference in accumulating errors in the 1-step model is directly reflected in the value function estimated with the model. Figure~\ref{fig:pgm-value} shows how a 1-step model leads to more errors in the distribution of returns compared to a VOC model. Figure~\ref{fig:rollouts} shows autoregressive rollouts produced by each of these models for 5 model timesteps. Interestingly, we notice that the one-step VOC model leads to slightly diverging predictions as compared to the $\gamma=0.8$ VOC model. The inferior accuracy of the 1-step model is also showcased in the return distributions presented in Figure~\ref{fig:pgm-value}.  This result highlights how TD-learning might be a superior objective when learning autoregressively over pixel spaces.

\vspace{0.5cm}
\begin{remark}
We use a TD(1) objective in VOCs, where the (1) corresponds to using the next representation and then following up by the bootstrap prediction. Similarly, we can use a multi-step TD objective, where a representation $k$ timesteps in the future is first sampled with probability $\gamma^{k-1} (1-\gamma)$ and then we sample the bootstrap prediction with the remaining probability. This forms a $k$-step TD estimator, akin to the case of n-step value estimation with rewards.  
\end{remark}

\begin{wrapfigure}{r}{0.6\textwidth}
    \centering
    % left bottom right top
    % \vspace{-1.7cm}
    \includegraphics[trim={1cm 0cm 1cm 0cm}, clip, width=1.\linewidth]{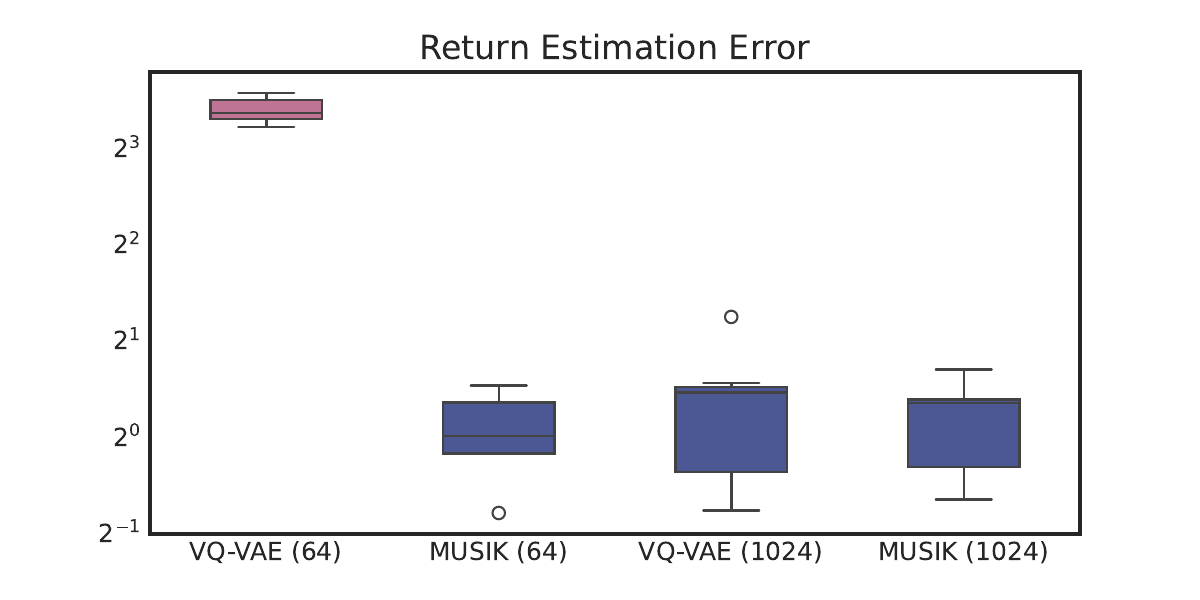}
    \caption{\textbf{Inverse Dynamics Modelling VOCs} with varying codebook sizes. We compare VQ-VAE and quantized MUSIK representations for different codebook sizes. Results are for the cheetah domain. Value estimation follows Eq.~\ref{eq:value-estimation-sample}. Standard codebook size is 1024.}
    \label{fig:pgm-musik}
\end{wrapfigure}

\newpage
\subsection{Inverse Dynamics Modelling}

VQ-VAE spaces preserve most pixel-level information present in the input. However, in the case when we solely care about control, a lot of the information can be dropped. Learning a multistep inverse kinematics (MUSIK)~\citep{efroni2021provably, lamb2022guaranteed, islam2022agent} model is one way to achieve a more compressed representation of the input. MUSIK works by learning to predict the action given the current and future observation in a trajectory. The conditioning over the future input enforces learning long-term dependencies within the representation so as to better predict the action. In this section, we develop a quantized version of the MUSIK representation. Figure~\ref{fig:pgm-musik} shows value estimation error plots when using MUSIK representations over VQ-VAEs. We are able to achieve a healthy reduction in codebook size when using more compact representations like MUSIK. This result showcases how most information can be successfully dropped (first through the MUSIK objective and then through the quantization) without effecting the downstream value estimates by much.

\subsection{Self-supervised (distillation) Modelling}

VQ-VAEs are based on convolutional networks (ConvNets) and have been shown to be less effective than transformer (ViT)-based self-supervised methods in capturing saliency maps of objects in images~\citep{caron2021emerging}. This raises a natural question about comparing VQ-VAEs with more advanced tokenization schemes. Specifically, we use quantized DINO features as the token space for VOCs. The quantized version of DINO incorporates a VQ-bottleneck immediately after the ViT encoder, enabling it to provide tokens for an image. 

\begin{wrapfigure}{r}{0.6\textwidth}
    \centering
    % left bottom right top
    \vspace{-.2cm}
    \includegraphics[trim={0cm 0cm 0cm 0cm}, clip, width=\linewidth]{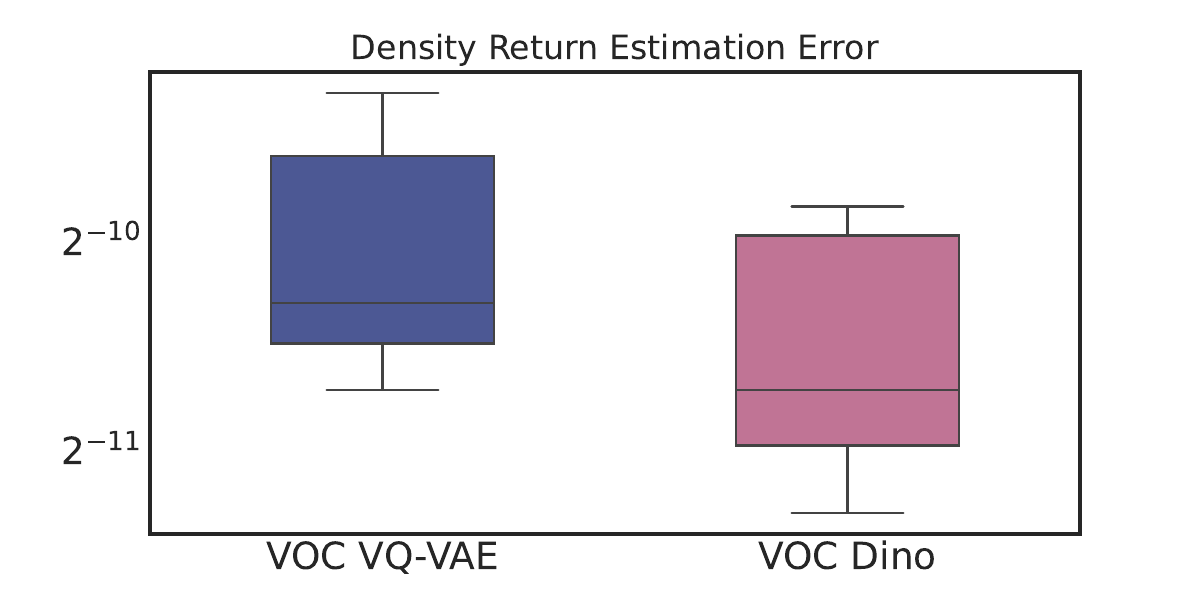}
    \caption{\textbf{VOCs with DINO Representations}. We compare VQ-VAE and quantized Dino based Video Occupancy Models. We compute the return estimation error by evaluating the density of a given trajectory under the two models. Both return estimates use the same reward estimate, i.e. that from the VQ representation space pre-discretization.}
    \label{fig:pgm-dino}
\end{wrapfigure}

We compare VQ-VAE and DINO-based Video Occupancy Models side by side in Figure~\ref{fig:pgm-dino}. Here we compare the return estimation error computed via density estimation (using Eq.~\ref{eq:value-estimation-density}) for both the VQ-VAE VOC and DINO VOC models. Since the density-based return estimator requires a reward model, we use a common reward model for both DINO and VQ-VAE representations to ensure a fair comparison. We observe that quantized DINO representations lead to improved density value estimates on validation trajectories, resulting in lower errors in the corresponding return estimates. To ensure fairness, we use a larger codebook size for the VQ-VAE model, so the GPT models used on top of VQ-VAE and DINO tokens have identical parameters.

% \begin{figure}[t] %{r}{0.5\textwidth}

\subsection{Control with Video Occupancy Models}

Having shown how VOCs can be learnt with different representation spaces and deployed for value function estimation, we finally move to using them for control. Our goal in this section is to use the most minimally feasible model-based setup to showcase VOCs ability to do downstream control. To that end, we adopt a model predictive control (MPC)~\citep{nagabandi2018neural} approach. MPC works by querying the model with different candidate actions and choosing the action that leads to the maximum reward under the model's predictions. Since VOCs are not action conditioned, we cannot use them as is for ranking the utility of actions as in standard MPC. Instead, we select a set of candidate actions first and use a simulator to provide next step observations corresponding to the candidate actions given a starting observation. We then query the model for value estimates for these set of next step observations. Finally, we choose the action that leads to the observation with the maximum value as predicted by the VOC model. 

\begin{figure}[t] %{r}{0.5\textwidth}
% \begin{wrapfigure}{r}{0.6\textwidth}
    \centering
    % left bottom right top
    \vspace{-.2cm}
    \includegraphics[trim={2cm 0cm 1.5cm 0cm}, clip, width=0.49\linewidth]{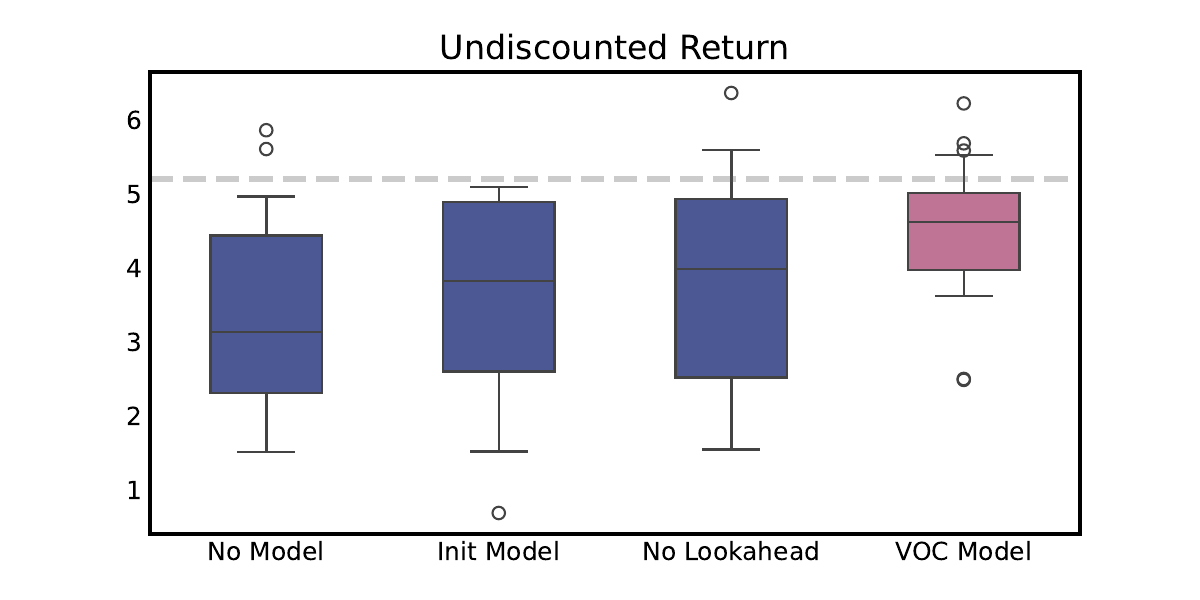}
    \includegraphics[trim={1.5cm 0cm 2cm 0cm}, clip, width=0.49\linewidth]{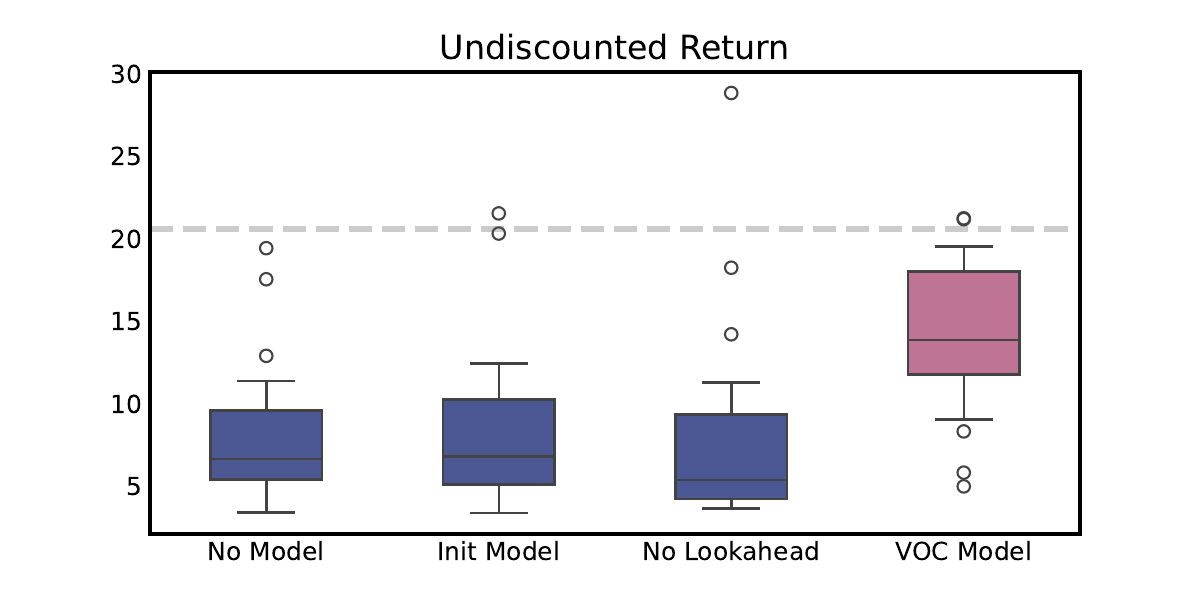}
    \caption{\textbf{Model Predictive Control with VOC Models}. Undiscounted return computed for episodes of length 20. Left plot is the cheetah and right plot is the walker environment. Each version is run 20 times. Candidate actions are selected from a buffer of expert actions of size 100.}
    \label{fig:pgm-control}
\end{figure}

Figure~\ref{fig:pgm-control} shows undiscounted return estimates across an episode length of 20 when using MPC with a VOC model. We include three natural baselines: 1) \textit{No Model}: where actions are randomly picked from the candidate action set, and 2) \textit{Init Model} where we pick actions according to value estimates generated from a randomly initialized GPT model used by the VOC, and 3) \textit{No Lookahead} model where we choose the best action that leads to the highest predicted reward for the achieved next state, i.e. discarding multi-step return information for single step reward. Note that for 2), the reward model still carries important information about the utility of the next observation. Therefore, this baseline helps isolate the benefit of learning the underlying GPT model when doing control. We observe higher returns for the VOC model than compared to the three baselines. Note that for the cheetah domain, there is a bit of an overlap in the box plots between all four methods since there is room for a randomly chosen action to perform well for the chosen horizon value. However, overall, VOC value estimates can steer the control policy towards high returns.

\section{Related Work}

\textbf{Video Predictive Models.} The work in learning video prediction models can be broadly categorized into two main approaches: learning representations from video frame sequences and developing generative models to produce future frames. 

In representation learning, the prominent methods include VideoMAE~\citep{tong2022videomae}, which reconstructs masked videos, extending the Masked Autoencoders (MAE) concept from static images to multiple frames. A related approach involves masked image modeling that predicts within a representation space~\citep{li2023mage}. This method has been adapted for videos, where the prediction target is derived from future frame patches~\citep{assran2023self}. Temporal contrastive learning approaches~\citep{sermanet2018time} have also been widely employed for learning control-oriented representations~\citep{nair2022r3m}. Despite leveraging temporal information for prediction, these methods primarily result in fixed representations of given observations, rather than models capable of performing rollouts.

% \end{wrapfigure}

On the generative front, diffusion techniques have been extensively utilized to generate video frames by corrupting an entire sequence and then predicting the original, uncorrupted frames~\citep{ho2022imagen}. Latent diffusion~\citep{rombach2022high} operates in a VQ-VAE latent space~\citep{van2017neural} rather than directly in the pixel space. VQ-VAE embeddings have also been integrated with various transformer architectures, such as TECO and VideoGPT~\citep{yan2021videogpt}. To our knowledge, these generative methods focus exclusively on pixel-level future frame generation and have not demonstrated efficacy for downstream control or value estimation. Moreover, these models typically generate predictions one step at a time, necessitating iterative rollouts to achieve long-term predictions.

\textbf{Successor Representation and Related Ideas.} The Successor Representation (SR)~\citep{dayan1993improving} is a well-established concept in cognitive neuroscience and reinforcement learning. SR aims to encapsulate a summary of the future states likely to be encountered, independent of a specific reward function. This predictive map can be learned through Temporal Difference (TD) learning, using indicator state occupancy rewards. In a deep learning context, SR has been utilized by incorporating user-specified features and learning an SR estimation of these features~\citep{kulkarni2016deep}. A more recent method, the FB representation~\citep{touati2021learning}, eliminates the need for user-specified features by imposing a linear structure on the SR through forward and backward representations. However, the FB representation has yet to be applied to pixel-based observation tasks. Another related concept involves learning intent-conditioned value functions (ICVF), which use an intent variable to learn value functions, similar to how the F and B representations are parameterized~\citep{ghosh2023reinforcement}. However, ICVF does not define intents over a representation space; instead, it uses resampled future frames as intents, encountering issues akin to the original indicator-reward based SR formulation. Lastly, $\gamma$-models~\citep{janner2020gamma} and $\delta$-models~\citep{wiltzer2024distributional} propose learning generative and distributional analogs of SR, respectively. These models aim to capture not just the expectation but the entire distribution of future states. Both approaches still rely on indicator-rewards and have not been applied to pixel-based observations.

\textbf{Representation Learning via Reinforcement Learning.} Recent research in control tasks often involves either pre-training representations through self-supervised learning or learning features in an end-to-end manner using a specific reward function. The general value function (GVF) principle~\citep{sutton2011horde} offers a promising alternative by learning features on-the-fly through a Temporal Difference (TD) paradigm. This approach involves making predictions about features via temporally bootstrapped target predictions, enabling feature learning without relying on a specific extrinsic reward function. TD learning is powerful because it refines predictions using backups of incomplete models. Despite its potential, GVFs typically struggle due to the absence of a clear utility function for the features being learned. We address this issue by using representations learned from self-supervised methods as proxies for evaluating the usefulness of certain features. This allows the temporal difference objective to continue shaping these features, thus leveraging both self-supervised learning benefits and the dynamic refinement capabilities of TD learning. 

\section{Limitations and Future Work}

This paper combined various representation spaces within VOC models. However, we restricted to pre-learnt representations for the most part. It would be interesting if the temporal predictions from the generative model can be used as a target to learn the VOC representation itself. 
Furthermore, this paper dealt with simple representation learning methods such as the VQ VAE and so required stacking VQ codes corresponding to multiple frame for temporal prediction. Such a design is non-ideal since the VQ codes contain a lot of redundant information. For instance, the background content between three consecutive frames remains almost the same. A downside of carrying over such redundant information is that temporal prediction is restricted to fewer timesteps in the future. Future work can alleviate this issue by using representation methods that capture information across multiple frames (e.g. video MAE representations), i.e. capture dyanmics-relevant information in the VOC representations. This would reduce redundant information and encourage predicting temporally for even longer horizons. Essentially, processing information across a stack of frames would then be amortized via both the representation and the temporal prediction axes. 

Finally, we presented one way to incorporate predictions made by VOCs for better control, i.e. via model predictive control. There remain multiple more powerful ways to combine VOC-like world models with model-based control, including search methods over the learned model~\citep{silver2018general}.

\section{Conclusion}

We introduce a new family of video prediction models called Video Occupancy Models (VOCs). VOCs work in a well defined latent space (that of discrete codes produced via different representation methods) and temporally predict via a generative TD loss. Since they are essentially trained with a generative version of the Successor Representation, they circumvent the issue of unrolling a model multiple times to produce future predictions. Instead, they produce future predictions (as governed by the parameter $\gamma$) within a single forward pass. We show that not predicting at every timestep leads to better and faster rollout accuracy. Furthermore, these models can be used to value estimation when given access to a reward function, and subsequently can be embedded in a model-based algorithm for downstream control. 

\section{Acknowledgments}

Part of this work has taken place in the Intelligent Robot Learning (IRL) Lab at the University of Alberta, which is supported in part by research grants from the Alberta Machine Intelligence Institute (Amii); a Canada CIFAR AI Chair, Amii; Compute Canada; Huawei; Mitacs; and NSERC.

\bibliography{bibliography}
\bibliographystyle{icml2024}

\newpage
\appendix

\section{Implementation Details}

\begin{center}
\begin{table}[ht]
\caption{\textbf{Hyperparameters for GPT Training.} These are the hyperparameters for training VOCs' underlying GPT model.}\label{tab:hyperparam_umd_pixel}
\def\arraystretch{1.35}
\begin{tabular}{|l|c|} 
\hline
\textbf{Model Variant} & GPT-2 \\
\textbf{Optimizer} & AdamW~\citep{loshchilov2017decoupled} \\
\textbf{LR}  & 3e-4 \\
\textbf{Batch Size}  & 32 \\
\textbf{Vocabulary/Codebook Size} & 1024 \\
\textbf{EMA for GPT Target $(M')$} & 0.9 \\
\textbf{Weight Decay}  & 0.05\\
\textbf{Optimizer Momentum} & $\beta_1, \beta_2$ = 0.9, 0.95\\
\textbf{Learning Rate Schedule}  & Cosine Decay~\citep{loshchilov2016sgdr}\\
\textbf{Warmup Steps} & 5000 \\
\textbf{Training Steps} & 100000 \\
\textbf{Augmentation} & Random Resized Crop(0.8, 1.0) \\
\hline
\end{tabular}
\end{table}
\end{center}

\begin{center}
\begin{table}[ht]
\caption{\textbf{Hyperparameters for VQ-VAE Training.} These are the hyperparameters for training/fine-tuning VOCs' underlying VQ-VAE representation model.}\label{tab:hyperparam_umd_pixel}
\def\arraystretch{1.35}
\begin{tabular}{|l|c|} 
\hline
\textbf{Model Variant} & VQ-GAN \\
\textbf{Optimizer} & AdamW~\citep{loshchilov2017decoupled} \\
\textbf{LR}  & 3e-4 \\
\textbf{Batch Size}  & 64 \\
\textbf{Pretrained}  & Yes, ImageNet \\
\textbf{Input Shape}  & 64 $\times$ 64 $\times$ 9 \\
\textbf{FrameStack}  & 3 \\
\textbf{Vocabulary/Codebook Size} & 1024 \\
\textbf{Optimizer Momentum} & $\beta_1, \beta_2$ = 0.9, 0.95\\
\textbf{Training Steps} & 10000 \\
\textbf{Augmentation} & Random Resized Crop(0.8, 1.0) \\
\hline
\end{tabular}
\end{table}
\end{center}

\textbf{Quantized DINO Training.} For DINO training, we use a pretrained BEiT-2~\citep{wang2022image} which includes a quantized ViT encoder trained to match the outputs of a pretrained DINO model. The BEiT encoder has a codebook size of 8192, much larger than the VQ-VAE representation spaces (which uses 1024 by default). The rest of the training details follow the same procedure as described in the Hyperparameter table containing GPT details.

\textbf{Quantized Inverse Dynamics Modelling}. For the MUSIK objective, we use the same VQGAN encoder as used by our VQ-VAE implementation, and replace the convolutional decoder with a 2-layer MLP head that predicts the action. For the input to the MUSIK model, we encode a current and future observation ($k$ steps in the future) using the VQGAN encoder, and concatenate the post-quantization embeddings of both the observations before passing into the MUSIK decoder. We randomly sample $k$ from 1 to 15. 

\end{document}